\documentclass[10pt,twocolumn,letterpaper]{article}

\usepackage{wacv}              

\usepackage{graphicx}
\usepackage{amsmath}
\usepackage{amssymb}
\usepackage{booktabs}
\usepackage{url}
\usepackage[pagebackref,breaklinks,colorlinks]{hyperref}
\usepackage[capitalize]{cleveref}
\crefname{section}{Sec.}{Secs.}
\Crefname{section}{Section}{Sections}
\Crefname{table}{Table}{Tables}
\crefname{table}{Tab.}{Tabs.}

\def\wacvPaperID{2660}
\def\confName{WACV}
\def\confYear{2025}

\begin{document}

\title{SEMU-Net: A Segmentation-based Corrector for Fabrication Process Variations of Nanophotonics with Microscopic Images}

\author{
    Rambod Azimi\textsuperscript{1}\textsuperscript{*}, Yijian Kong\textsuperscript{1}\textsuperscript{*}\textsuperscript{†}\\
    Dusan Gostimirovic\textsuperscript{1}, James J. Clark\textsuperscript{1}, and Odile Liboiron-Ladouceur\textsuperscript{1} \\
    \textsuperscript{1}{McGill University, Canada} \\
}

\maketitle

\renewcommand{\thefootnote}{\fnsymbol{footnote}}
\footnotetext[1]{Equal contribution.}
\footnotetext[2]{Corresponding author.}

\begin{figure*}[t]
   \centering
    \includegraphics[width=1\linewidth]{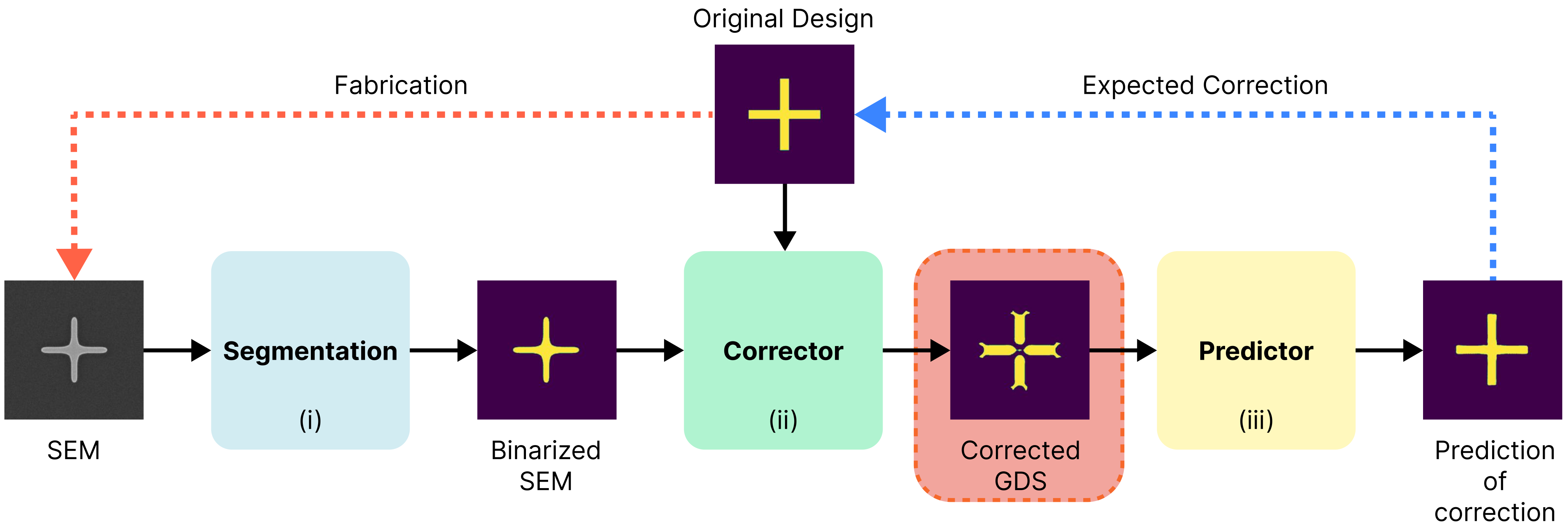}
    \caption{Overview of the SEMU-Net framework for improving the fabrication of integrated silicon photonic devices. (i) The segmentation model converts SEM images into segmented SEM images. (ii) The corrector model uses the binarized SEM images along with their corresponding GDS design files to train itself, generating corrected GDS layouts that compensate for fabrication-induced variations. (iii) The predictor model uses the corrected GDS files to predict the final fabricated structures, enabling pre-fabrication validation and further refinement.}
    \label{semu-net framework}
\end{figure*}

\begin{abstract}
Integrated silicon photonic devices, which manipulate light to transmit and process information on a silicon-on-insulator chip, are highly sensitive to structural variations. Minor deviations during nanofabrication—the precise process of building structures at the nanometer scale—such as over- or under-etching, corner rounding, and unintended defects, can significantly impact performance. To address these challenges, we introduce SEMU-Net, a comprehensive set of methods that automatically segments scanning electron microscope (SEM) images and uses them to train two deep neural network models based on U-Net and its variants. The predictor model anticipates fabrication-induced variations, while the corrector model adjusts the design to address these issues, ensuring that the final fabricated structures closely align with the intended specifications. Experimental results show that the segmentation U-Net reaches an average IoU score of 99.30\%, while the corrector attention U-Net in a tandem architecture achieves an average IoU score of 98.67\%.
\end{abstract}

\section{Introduction}

\begin{figure}[t]
   \centering
    \includegraphics[width=1.0\linewidth]{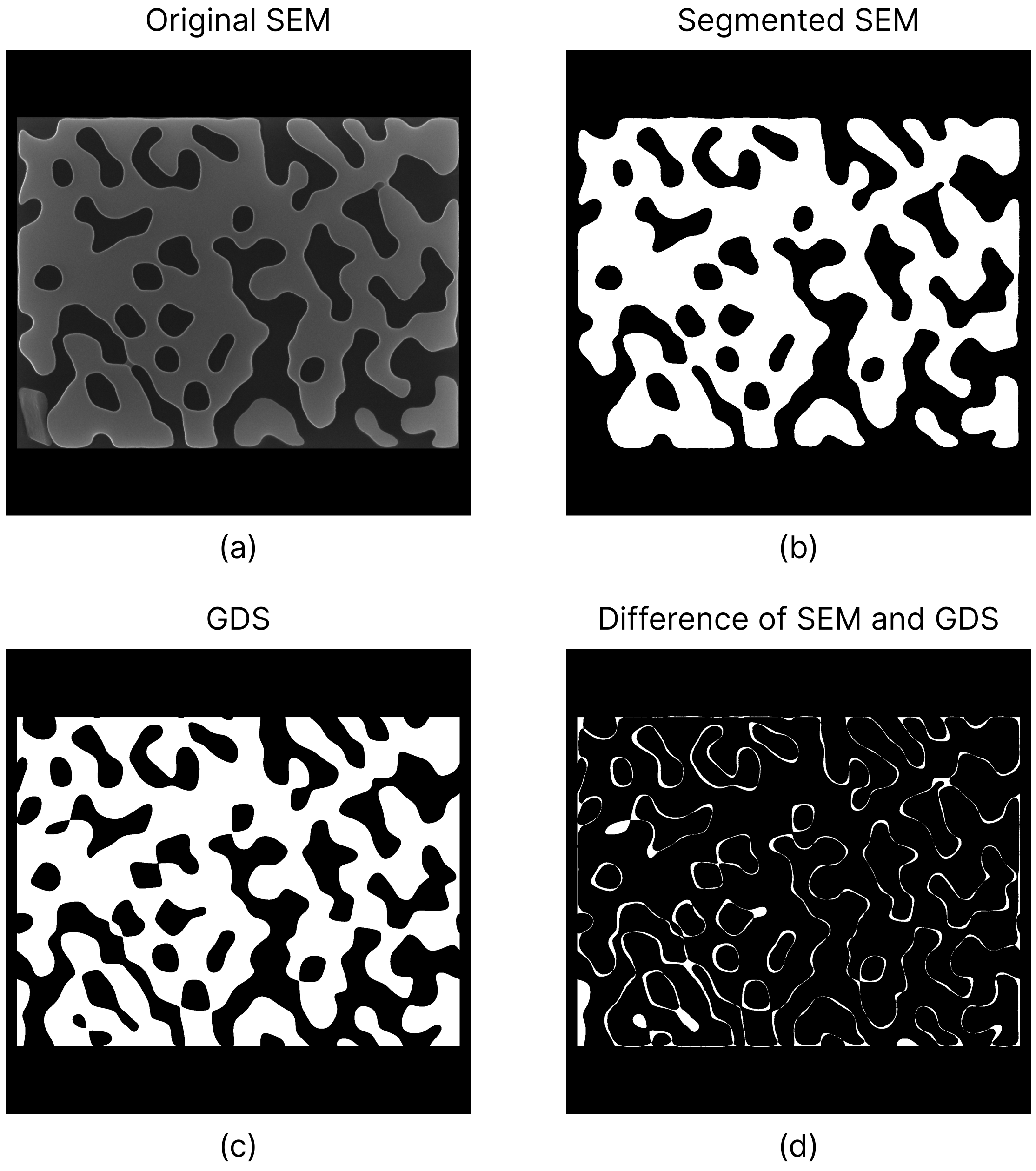}
    \caption{A sample from the ANT NanoSOI dataset: (a) an SEM image of a photonics structure with silicon (gray) and silica (black), (b) its segmented SEM image, (c) its corresponding GDS design file, and (d) the difference between the segmented SEM image and the GDS design file.}
    \label{sample dataset}
\end{figure}

Integrated silicon photonic devices operate using light (photons) \cite{dong2014siliconphotonicdevicesandintegratedcircuits} rather than electrons to perform various functions through a silicon waveguide on a silicon-on-insulator (SOI) die. This approach offers significant advantages over traditional electronic devices \cite{AMSALU20203372}. Photonic devices can achieve lower latency, reduced heat generation, and minimal energy loss, making them ideal for high-bandwidth data transmission and processing tasks \cite{acsphotonics, ning2024photonicelectronicintegratedcircuitshighperformance}. This efficiency is crucial given the growing computational demands of AI and machine learning systems \cite{reviewofsiliconphotonics}, which often require significant resources for managing and processing complex models\cite{brown2020languagemodelsfewshotlearners, chowdhery2022palmscalinglanguagemodeling}.

Integrated silicon photonic devices often underperform experimentally due to fabrication-induced structural variations \cite{acsphotonics2}. Nanometer-scale deviations, such as over-etching, under-etching, and corner rounding, can significantly degrade device performance, efficiency, and functionality \cite{acsphotonics}. Research indicates that these variations tend to follow consistent patterns, which may allow them to be learned and mitigated \cite{acsphotonics}.

With advancements in AI and particularly convolutional neural networks (CNNs) \cite{oshea2015introductionconvolutionalneuralnetworks}, researchers have achieved remarkable improvements in computer vision tasks, including image \cite{wu2015deepimagescalingimage, dosovitskiy2021imageworth16x16words} and video recognition \cite{zhu2020comprehensivestudydeepvideo}. These sophisticated architectures have enabled the development of models that can automatically learn and extract features from visual data, enhancing accuracy, efficiency, and robustness in tasks such as object detection \cite{Sultana_2020} and semantic segmentation \cite{ronneberger2015unetconvolutionalnetworksbiomedical}.

In this paper, we introduce SEMU-Net (see Figure \ref{semu-net framework}), a comprehensive set of methods designed to automate the segmentation of scanning electron microscope (SEM) images of integrated silicon photonic devices into binary classifications of silicon (core waveguide confining light) and silica (cladding material around the core). These segmented images, along with their corresponding design files in the Graphic Data System (GDS) format—a standard file format used in the photonics industry for representing the physical layout of devices—are then used to train two CNN models: the predictor and the corrector.

The segmentation model is designed to automatically segment SEM images of photonic devices. It takes SEM images as input and uses manually segmented images as labels, leveraging a traditional U-Net\cite{ronneberger2015unetconvolutionalnetworksbiomedical} architecture to achieve accurate segmentation of previously unseen SEM images. Figure \ref{sample dataset} presents a segmented SEM image of a fabricated structure, along with its original SEM image, the corresponding GDS design file, and the difference between the SEM of the fabricated structure and the intended design in the GDS layout.

The predictor model learns to map the actual GDS design file (the intended designed structure) to the SEM image (the fabricated device), thereby identifying structural variations that occur during fabrication and enabling the prediction of potential issues before fabrication begins. This predictive capability can significantly reduce both cost and time. Conversely, the corrector model uses an inverse design approach \cite{acsphotonics3} to map SEM images to GDS design files, enabling the adjustment of design files to better match the desired output. By employing this reverse approach, the corrector model generates design files with deliberate exaggerations in areas identified by the predictor model as prone to fabrication changes. As a result, the corrected design files more accurately align with the intended design after fabrication.

The SEMU-Net integrates three models—segmentation, predictor, and corrector—into a unified framework, streamlining the entire process within a cohesive system.

For the model architecture, we employ U-Net \cite{ronneberger2015unetconvolutionalnetworksbiomedical}, a widely used encoder-decoder-based CNN, to train our models using the ANT NanoSOI dataset, provided by Applied Nanotools Inc \textsuperscript{1}. The U-Net architecture is well-suited for this application due to its effectiveness in segmenting complex structures and handling the intricate details present in the photonic circuit images. In addition to the standard U-Net, we explore various U-Net variants, including attention U-Net \cite{oktay2018attentionunetlearninglook}, residual attention U-Net \cite{ni2019raunetresidualattentionunet}, and U-Net++ \cite{zhou2018unetnestedunetarchitecture}, to determine the optimal performance while maintaining relatively low computational costs.
\renewcommand{\thefootnote}{\arabic{footnote}}
\footnotetext[1]{\url{https://www.appliednt.com/nanosoi-fabrication-service}}

To enhance the results further, we employ a tandem architecture by stacking the predictor and corrector models, freezing the weights of the predictor, and updating the corrector to learn the identity mapping from GDS to GDS. We assess the model's effectiveness using the Intersection-over-Union (IoU) score. The experimental results reveal that the segmentation U-Net achieves an average IoU score of 99.30\%, while the corrector model attains an average IoU score of 98.67\%, evaluated on a custom benchmark consisting of various distinct shapes, including gratings, stars, crosses, and circles. The segmentation model is tested using the traditional U-Net with a binary cross-entropy (BCE) loss function, while the corrector model is tested using the attention U-Net in the tandem configuration, with a loss function that combines BCE and a weighted 0.5 dice loss.

\section{Related Work}
\subsection{Segmentation of SEM Images}

Segmenting SEM images presents unique challenges due to the high level of noise, varying contrast, and intricate textures present in these images. Traditional image processing techniques, such as thresholding \cite{alamri2010imagesegmentationusingthreshold} or edge detection \cite{cannyedgedetection}, often fall short of accurately segmenting these images.

In recent years, deep learning models like U-Net have been applied to image segmentation tasks with promising results. For instance, the U-Net architecture, originally introduced by \cite{ronneberger2015unetconvolutionalnetworksbiomedical}, has become a standard in the field of image segmentation, particularly for biomedical images. Its encoder-decoder structure allows it to capture context while maintaining spatial resolution, making it effective for segmenting images with complex and fine details. The U-Net model's strength lies in its ability to work with limited datasets and its flexibility to adapt to various segmentation tasks, which has led to its adoption in different domains, including SEM image analysis.

Various enhancements to the U-Net architecture have been proposed to improve its performance in specific tasks. For example, \cite{yu2015multi} introduced dilated convolutions to increase the receptive field without losing resolution, which is useful in detecting fine structures in high-resolution images. Other variants, such as the attention U-Net \cite{oktay2018attentionunetlearninglook}, have incorporated attention mechanisms to focus on the most relevant features, which could be beneficial in handling the high noise levels often present in SEM images.

While there has been significant progress in using deep learning for SEM image segmentation, existing works have primarily concentrated on applications outside the photonics domain or on general image segmentation tasks. Our work, therefore, applies a machine learning model specifically for segmenting SEM images of nanophotonics. This novel application addresses the unique requirements of photonic integrated structures, which involves specific structural features and high-precision segmentation not covered by prior deep learning methods.

\subsection{Image Translations for Photonic Devices}

Photonics devices, which rely heavily on precise geometrical configurations, can undergo shape changes due to fabrication imperfections from inherent process variations. Predicting these shape changes is crucial for maintaining device performance and reliability. Traditional methods for predicting shape changes in photonics devices include finite element analysis and other simulation-based approaches in which features are explicitly biased to mimic over- and under-etching\cite{lu2017performance}. These methods do not capture, however, the process variations leading to corner rounding and curvature changes. In contrast, SEM images, which can serve as training data, are more readily available, providing an alternative pathway for predicting shape changes allowing to capture a greater range of fabrication process variation.

There has been considerable research in using machine learning techniques for image-to-image translation, with applications spanning various domains. Notable works in this area include the development of generative adversarial networks (GANs) for tasks such as translating sketches to photos \cite{isola2017image}, converting day images to night images \cite{zhu2017unpaired}, and performing style transfer \cite{gatys2016image}. These studies highlight the potential of machine learning to transform images across domains while preserving core content.

When it comes to photonics, the application of machine learning for predicting and correcting structural changes is relatively underexplored. For example, our recent work presented in \cite{acsphotonics} laid the groundwork for integrating machine learning methods towards better performing photonic integrated devices. However, there remains significant room for further exploration and enhancement, particularly in improving prediction accuracy and generalizability to greater photonic structures.

Our approach addresses this gap by integrating SEM image segmentation with a predictive model that forecasts shape changes in photonic devices, along with a corrector model that refines the structure to maintain alignment with the original image after structural variations. This integration not only enhances the accuracy of shape predictions but also streamlines the overall process, making it more applicable to real-world scenarios.

\subsection{Summary of Gaps and Contributions}

In summary, while there has been significant progress in the image segmentation domain and the application of machine learning in photonics, key gaps remain. Most notably, the lack of integrated tools for segmentation and dynamic shape prediction in photonic integrated devices presents a challenge for practical applications. In addition, existing methods for predicting structural changes in photonic devices are computationally demanding, limiting their real-time use. Our work addresses these gaps by developing SEMU-Net, which integrates segmentation with predictive and corrective modeling to address shape changes in photonics devices.

\section{Methodology}

\begin{figure*}[t]
   \centering
    \includegraphics[width=1\linewidth]{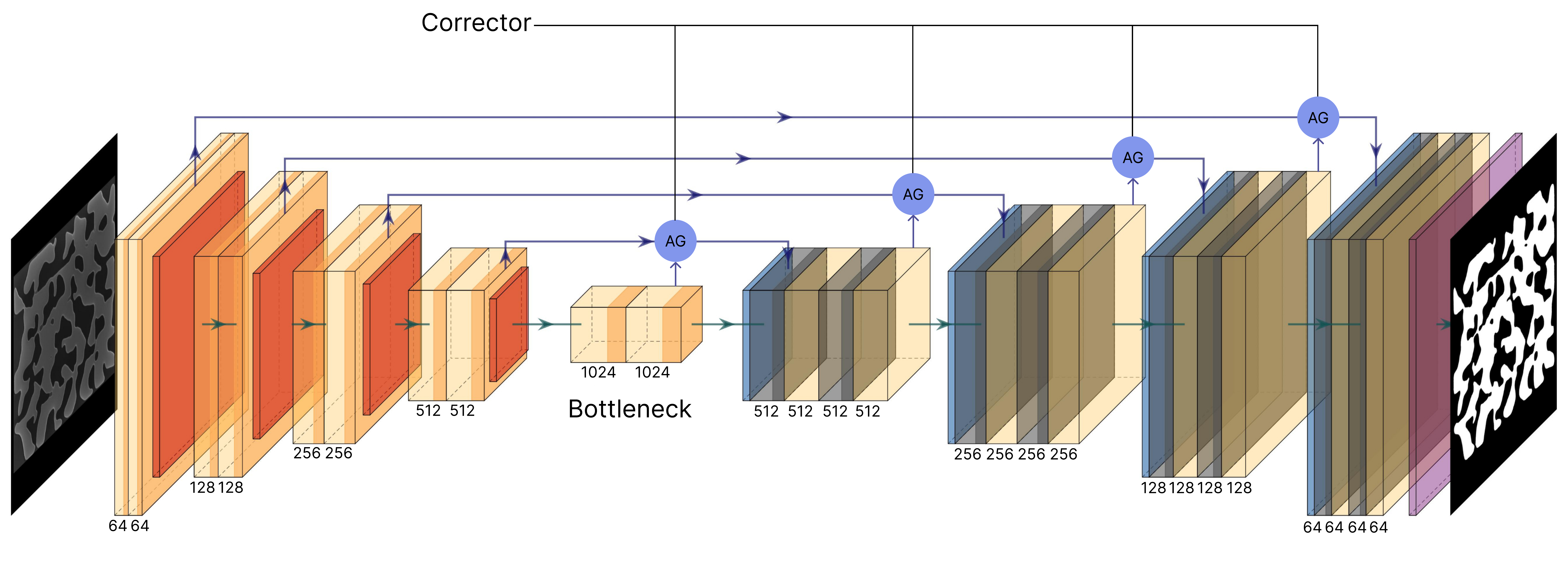}
    \caption{Overview of the segmentation, predictor, and corrector models, all based on the U-Net architecture. The corrector model distinguishes itself by incorporating attention gates (labeled AG) in the decoder path, whereas the segmentation and predictor models do not utilize attention gates. The segmentation model takes SEM images as input and generates segmented labels, while the corrector model takes GDS designs as input and outputs corrected designs.}
    \label{model architecture}
\end{figure*}

In this section, we introduce the SEMU-Net, a framework for photonic design correction based on SEM images. As shown in Figure \ref{semu-net framework}, our model comprises three key components: the segmentation model for automatic label generation, the predictor model for forecasting structural variations during fabrication, and the corrector model for design rectification. The predictor model specifically evaluates the corrected version of a device image by applying predictions to the corrected design and comparing the resulting output with the original design file. This approach ensures that the corrections are accurately aligned with the intended design specifications.

These three modules are trained and run separately. We first explain our model structure and its corresponding principles, then introduce our dataset processing strategies, and finally, the training techniques.

\subsection{The Segmentation Model}
We use the original U-Net model, introduced by \cite{ronneberger2015unetconvolutionalnetworksbiomedical}, which is widely regarded for its effectiveness in image segmentation by capturing both fine details and broader contextual information. Our U-Net architecture, illustrated in Figure \ref{model architecture}, is structured into three main components: the encoder, bottleneck, and decoder paths. The encoder, or contracting path, uses a series of 3\(\times\)3 convolutional layers with ReLU activations followed by 2\(\times\)2 max-pooling layers to capture increasingly abstract features. At the network's base, the bottleneck layer further processes these features and links the encoder to the decoder. The decoder, or expansive path, uses upsampling and convolutional layers to reconstruct the segmentation map, with skip connections that concatenate feature maps from the encoder to preserve spatial details.

\subsection{The Predictor and Corrector Models}
\textbf{Predictor}: For predicting structural variations in photonic devices, we employ the same U-Net model as used in the segmentation U-Net, retaining the identical encoder and decoder paths. The only difference lies in the mapping images. In the segmentation U-Net, the model learns to map the original SEM image to its binary truth label. However, the predictor model learns to map the actual GDS design file to its corresponding segmented SEM image.

\textbf{Corrector}: Instead of mapping GDS to SEM images, the corrector model performs the reverse of the predictor model by learning to map SEM images back to GDS files. This enables the corrector to understand how to transition from a fabricated design to the original design, allowing it to adjust new design files so that the final fabricated result closely matches the intended design. For the correction of photonic devices, we employ the attention U-Net, first introduced by \cite{oktay2018attentionunetlearninglook}, which enhances the standard U-Net architecture with attention mechanisms to focus on relevant features.

\textbf{Attention Gate}: The attention block refines feature maps through the following steps: it first applies a 1\(\times\)1 convolution to reduce dimensions and then combines this with another 1\(\times\)1 convolution of the gating input. The resulting features are processed with a ReLU activation and another 1\(\times\)1 convolution to produce an attention map. This map, after being upsampled, is used to weight the original feature maps, enhancing important regions.

\subsection{Dataset Processing}
Our dataset pre-processing involves several key techniques to enhance the performance of the U-Net model.

\textbf{Data Augmentation}: To increase the robustness and generalizability of the model, we apply data augmentation techniques to the training images. The augmentation process is performed three times on each image with a 50\% probability for each transformation. The augmentations applied include horizontal flip, rotation, shift, scale, Gaussian noise, brightness change and contrast adjustment.

\textbf{Image Patching}: After applying data augmentation, the images are sliced into smaller patches to facilitate model training. Each image is divided into patches of size 256\(\times\)256 pixels. This patching process helps manage memory usage and allows the model to learn from localized regions of the images.

\textbf{Dataset Shuffling}: By randomly reordering the data samples before each training epoch, dataset shuffling helps prevent the model from learning unintended patterns based on the sequence of the data. In our approach, we employ dataset shuffling to reduce the risk of overfitting and enhance the model's generalization ability.

\subsection{Training Techniques}
\textbf{Tandem Architecture}: The tandem architecture is designed to refine the model by updating the weights during the correction stage, ensuring more accurate and reliable corrections. It sequentially integrates the correction and prediction processes to align the final output with the intended design. This architecture consists of two key components: the tandem corrector and the tandem predictor, stacked together. The tandem corrector refines the device structure, generating a corrected design file, while the tandem predictor evaluates this corrected file, simulating its post-fabrication appearance. A loss function is then calculated by comparing the predicted output to the original SEM image, with the goal of minimizing this loss to ensure the corrected output closely matches the SEM image. The tandem predictor model's weights are frozen to maintain consistency, while the corrector model's weights are updated at each iteration to improve the correction process.

\textbf{Early Stopping}: Early stopping is a technique used during model training to prevent overfitting and improve efficiency \cite{bai2021understandingimprovingearlystopping}. We use this technique by monitoring the validation loss and halting training when no significant improvement is observed after a certain number of iterations.

\textbf{Learning Rate Scheduler}: The learning rate scheduler is a method used to adjust the learning rate during training to improve model convergence \cite{defazio2023whenmuchadaptivelearning}. By gradually reducing the learning rate over time, the model can make finer adjustments as it approaches an optimal solution, helping to avoid overshooting minima in the loss function. We employ this technique to enhance training efficiency.

\textbf{Combined Loss Function}: The combination of BCE and 0.5 dice loss is used as a composite loss function to improve the correction performance \cite{Jadon_2020}. BCE focuses on pixel-wise classification, penalizing incorrect predictions for each pixel, making it effective for distinguishing between foreground and background. Dice Loss, on the other hand, measures the overlap between the predicted and ground truth masks, emphasizing the overall shape accuracy. By combining BCE with 0.5 dice loss, the model benefits from both fine-grained pixel accuracy and robust shape matching, leading to more precise segmentation results.

\textbf{Hyperparameter Tuning}: Hyperparameter tuning is crucial in optimizing model performance, involving adjusting parameters that control the learning process, such as learning rate, batch size, and network architecture \cite{yu2020hyperparameteroptimizationreviewalgorithms}. In our approach, we conduct a thorough search to optimize the hyperparameter set for all three models in our study. This process ensures that each model is configured with proper parameters to enhance overall performance and accuracy.

\section{Experiments}

\begin{figure*}[t]
   \centering
    \includegraphics[width=0.9\linewidth]{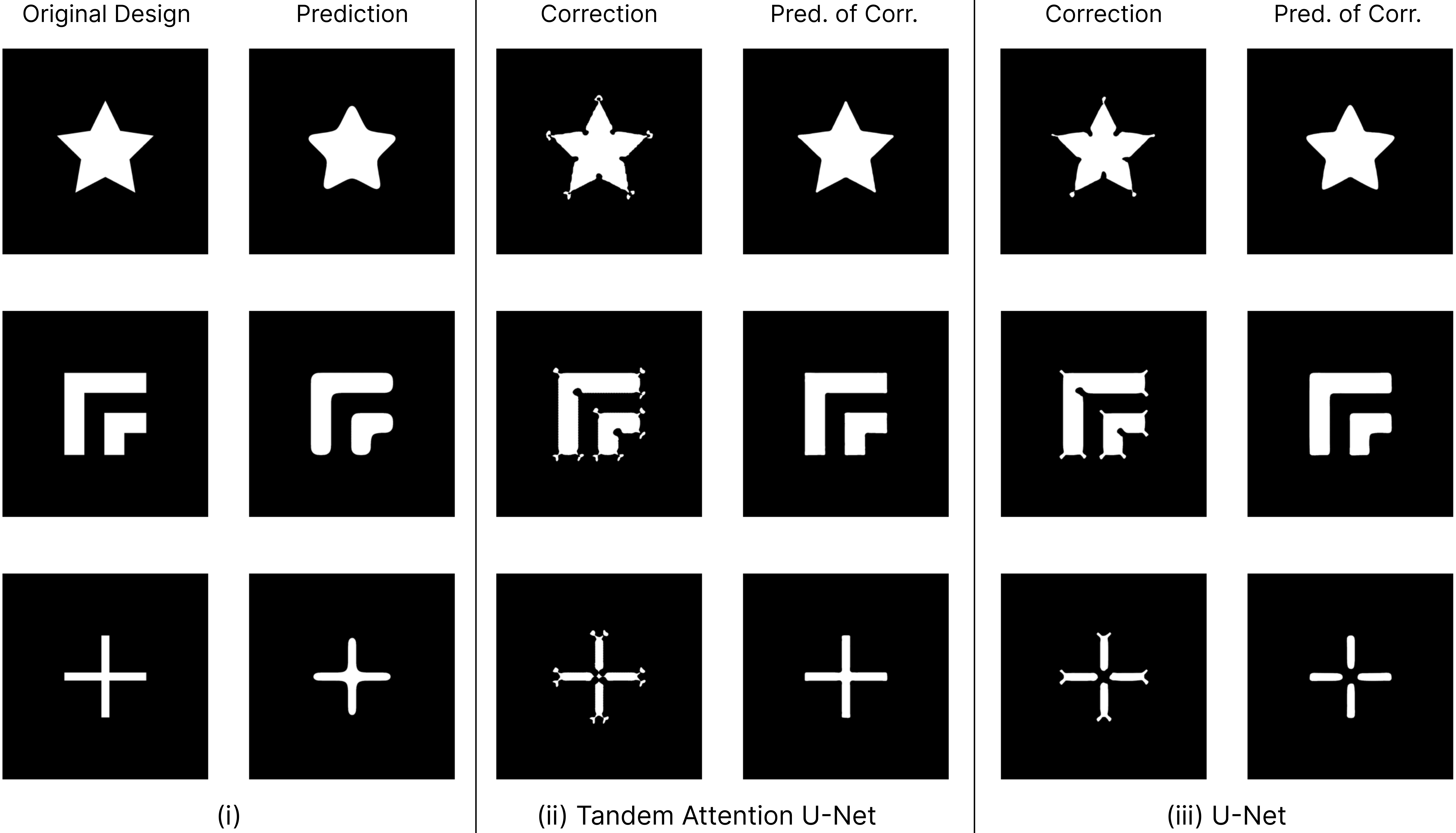}
    \caption{Performance comparison between the attention U-Net in tandem configuration and the original U-Net across three sample images. Each sample presents (i) the original design and its predicted structure post-fabrication, highlighting corner rounding and structural deviations that could impact device performance, (ii) the correction and prediction of correction for the tandem attention U-Net, and (iii) the correction and prediction of correction for the original U-Net. The tandem attention U-Net outperforms the original U-Net, demonstrating better structural fidelity and achieving a higher IoU score.}
    \label{corrector unet}
\end{figure*}

In this section, we evaluate each component of our proposed SEMU-Net framework. We present quantitative results for both the segmentation and correction models, comparing their performance using IoU as the metric. To rigorously assess the corrector model, we develop a custom benchmark comprising several hundred structure images featuring various shapes, including stars, gratings, circles, holes, and more (see Figure \ref{corrector unet}).

\subsection{Experiment Setup}
\textbf{Dataset}: For our dataset, we utilize Applied Nanotools Inc., a reputable integrated photonics foundry, for nanofabrication and SEM imaging to acquire high-quality data. Due to the high costs and lengthy timelines associated with nanofabrication, our training structures are designed to acquire a large, varied dataset with minimal chip space and imaging time.

The generated patterns are fabricated on a 220\(nm\) thick silicon-on-insulator (SOI) platform using electron-beam lithography through a silicon photonic multi-project wafer service provided by Applied Nanotools Inc. After lithography and etching, SEM images with a resolution of 1\(nm/pixel\) are taken of each pattern. The GDS and SEM images are then segmented with the help of our segmentation model, cropped, aligned, and prepared for training of the predictor and corrector models.

Finally, we augment the binarized SEM and GDS images to artificially increase their size and use these enhanced images to train our predictor and corrector models.

\textbf{Computational Resources}: The training and testing of our models are conducted using one GeForce RTX 4090 GPU, which is known for its good performance in deep learning tasks, and provides the necessary computational power to efficiently handle large datasets and complex operations involved in our SEMU-Net framework. This GPU allows for faster training times and enables us to experiment with various model configurations, ensuring robust and accurate results.

\textbf{Evaluation Metrics}: To evaluate the performance of our models, we use IoU as the primary metric, a widely used metric in segmentation tasks, providing a measure of the overlap between the predicted segmentation and the ground truth. In our experiments, we calculate the IoU for each pixel classification, determining whether it has been correctly categorized into silicon or silicon dioxide. A higher IoU score indicates better model performance, with an IoU of 1.0 representing perfect segmentation.

\textbf{Training Configurations}: The segmentation model is trained using image slices of size 256\(\times\)256 over 50 epochs, with a batch size of 32 to ensure robust evaluation. The model uses BCE as the loss function and a learning rate of 0.0001. On the other hand, the predictor and corrector models are trained with a larger slice size of 2048\(\times\)2048 over 20 epochs, employing early stopping after 3 epochs if no improvement in validation loss is observed. A batch size of 2 and a validation split of 20\% are utilized, with a learning rate of 0.0004 and the AdamW optimizer to ensure efficient training. Additionally, data augmentation is applied to enhance dataset diversity. These models are trained using a combination of BCE and 0.5 Dice Loss. For the tandem corrector model, the original training configurations are maintained, but the number of epochs is extended from 20 to 60 for more comprehensive learning.

\subsection{The Segmentation Model}
In this section, we compare the performance of our segmentation U-Net model with two baseline models: the segment anything model (SAM) \cite{kirillov2023segment} fine-tuned on our dataset, and a traditional threshold-based model. The goal is to evaluate the effectiveness of each model in segmenting SEM images of photonics devices.

\begin{figure*}[t]
   \centering
    \includegraphics[width=0.9\linewidth]{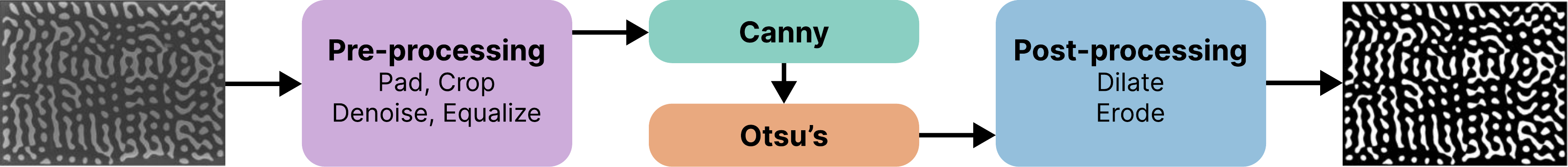}
    \caption{Workflow of the threshold-based segmentation model. A combination of Canny and Otsu's methods is used to detect the contours and further segment the pre-processed SEM images.}
    \label{thresmodel}
\end{figure*}

\textbf{Baseline Models}: (1) The attention U-Net extends the traditional U-Net architecture by incorporating attention mechanisms into the skip connections between the encoder and decoder. These attention gates focus on relevant features from the encoder while suppressing irrelevant information. The attention U-Net improves segmentation accuracy by dynamically highlighting important regions in the feature maps, which helps in handling more complex and varied structures. (2) SAM is a large-scale image segmentation model developed by Meta is based on vision transformer (ViT) architecture and is pre-trained on a vast dataset. For our experiments, we chose the heavy model ViT-H, which is then fine-tuned on our dataset to adapt its general segmentation capabilities to the specific characteristics of SEM images. This fine-tuning involves updating the model parameters to improve performance on our particular task. (3) The traditional threshold-based model is implemented following the workflow shown in Figure \ref{thresmodel}. This simple method is used as a baseline to evaluate the improvements achieved by more advanced models like U-Net and SAM. Despite its simplicity, this model provides a useful comparison point for assessing the effectiveness of more sophisticated approaches.

\begin{table*}[t]
   \centering
   \setlength{\tabcolsep}{6pt}
   \small
   \begin{tabular}{@{}lcccc@{}}
     \toprule
     Model & \# Parameters (M) & Average IoU (\%) & Max IoU (\%) & Min IoU (\%) \\
     \midrule
     \textbf{U-Net} & 7.94 & \textbf{99.30} & \textbf{99.71} & \textbf{98.35} \\
     SAM (ViT-H)  & 636 & 88.35 & 96.90 & 43.48 \\
     Threshold Model & NA & 96.54 & 98.59 & 86.16 \\ 
     \bottomrule
   \end{tabular}
   \caption{Comparison of the segmentation model performance on the custom benchmark. To ensure consistency, each experiment is run five times due to random dataset ordering and weight initializations, and the median result is taken.}
   \label{tab:seg result}
\end{table*}

\begin{figure*}[t]
   \centering
    \includegraphics[width=1\linewidth]{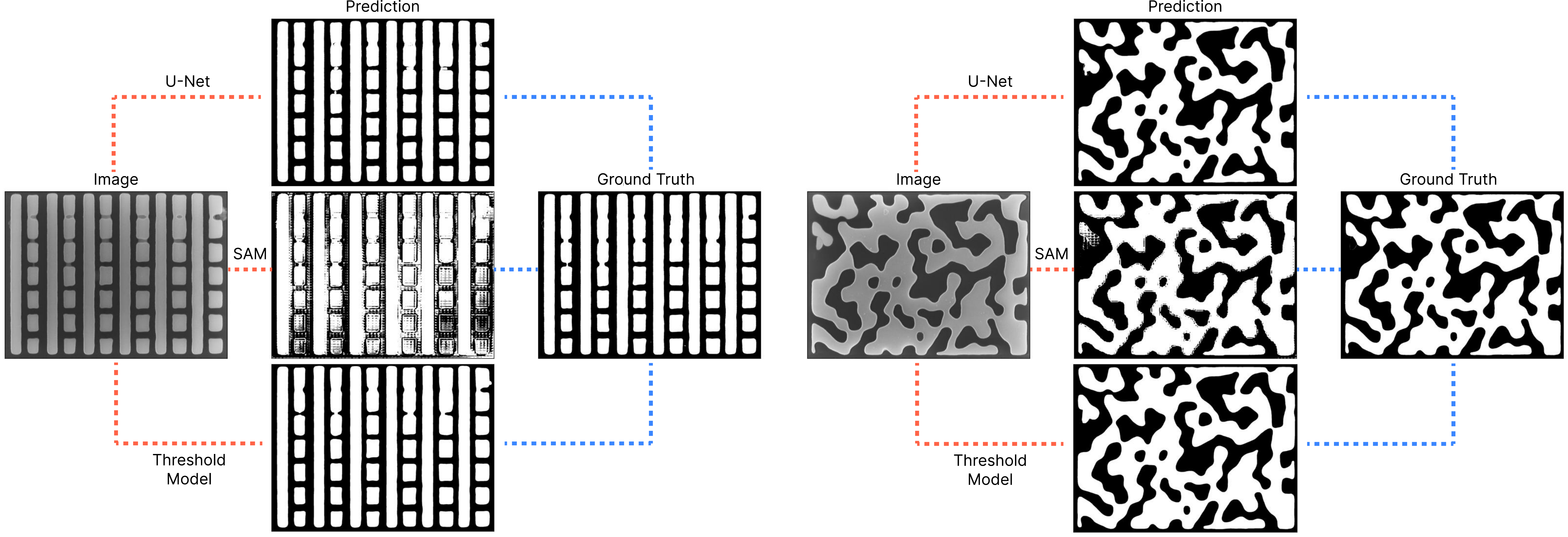}
    \caption{Performance comparison between the U-Net, SAM and threshold model for two sample SEM images, which are processed through the U-Net, SAM and threshold models and the segmentation results are compared with corresponding ground truth.}
    \label{seg result}
\end{figure*}

\textbf{Results}: The results of the comparison between the U-Net, SAM, and the traditional threshold-based model are summarized in Table \ref{tab:seg result}. The table shows the IoU scores for each model, providing their segmentation performance by the average IoU, maximum IoU and minimum of IoU, which reveals that the U-Net model outperforms all other models, achieving the highest IoU score. The threshold-based method comes in second and finally the SAM model. Two example SEM images are used to visualize the segmentation results from different models in Figure \ref{seg result}.

Despite its state-of-the-art status and extensive pre-training on a large dataset, SAM's performance in this context is relatively lower. SAM struggles to achieve optimal results due to its unfamiliarity with the specific segmentation task of SEM images. The model's generalization capabilities are limited when applied to a highly specialized dataset, indicating that fine-tuning SAM alone is insufficient for achieving top performance in this domain.

The U-Net's superior performance can be attributed to its straightforward architecture and effectiveness in handling complex segmentation tasks. Its symmetric encoder-decoder structure with skip connections allows it to capture both high-level contextual information and fine-grained details. This high accuracy demonstrates the model's robustness and suitability for SEM image segmentation, where precise delineation of features is crucial.

\subsection{The Predictor and Corrector Models}
In this section, we evaluate the performance of our corrector model using four different architectures: U-Net \cite{ronneberger2015unetconvolutionalnetworksbiomedical}, attention U-Net \cite{oktay2018attentionunetlearninglook}, residual attention U-Net \cite{ni2019raunetresidualattentionunet}, and U-Net++ \cite{zhou2018unetnestedunetarchitecture}. Additionally, we incorporate the tandem architecture into each model and assess the performance of all configurations. The aim is to assess the effectiveness of each model in correcting GDS images of photonics devices and to identify the most effective configuration for optimal performance.

\begin{table*}[!t]
   \centering
   \setlength{\tabcolsep}{6pt}
   \small
   \begin{tabular}{@{}lcccc@{}}
     \toprule
     Model & \# Parameters (M) & Average IoU (\%) & Min IoU (\%) & Median IoU (\%) \\
     \midrule
     U-Net & 1.94 & 92.38 & 6.12 & 97.01 \\
     Attention U-Net & 2.01 & 93.55 & 72.15 & 97.62 \\
     Residual Attention U-Net & 2.64 & 93.64 & 46.73 & 96.42 \\
     U-Net++ & 0.56 & 92.63 & 66.64 & 95.19 \\
     U-Net (tandem) & 3.88 & 97.87 & 71.39 & 98.90 \\
     \textbf{Attention U-Net (tandem)} & 4.02 & \textbf{98.67} & \textbf{88.86} & \textbf{99.51} \\
     Residual Attention U-Net (tandem) & 5.28 & 96.74 & 65.91 & 98.64 \\
     U-Net++ (tandem) & 1.13 & 93.98 & 58.55 & 96.58 \\
     \bottomrule
   \end{tabular}
   \caption{Comparison of the corrector model performance on the custom benchmark. To ensure consistency, each experiment is run five times due to random dataset ordering and weight initializations, and the median result is taken.}
   \label{tab:corrector results}
\end{table*}

\textbf{Results}: The results of the comparison between the U-Net, attention U-Net, residual attention U-Net, and U-Net++ and their tandem configuration are summarized in Table \ref{tab:corrector results}. The table presents the IoU scores for each model, providing a clear view of their correction performance by the average IoU, minimum IoU and the median of Iou, which reveal that the tandem attention U-Net model outperforms all other models, achieving the highest IoU score. To ensure consistency in the presence of random dataset ordering and weight initializations, each experiment is run five times, and the median of the results is used.

Processing images of size 2048\(\times\)2048 requires a substantial amount of GPU memory, which can be challenging to manage. To address this constraint, we reduce the filter sizes in our models, which, in turn, decreases the number of parameters and allows us to train the models efficiently within the limits of our available GPU resources.

The results indicate that while the base models show robust performance, the tandem versions, particularly the attention U-Net (tandem), outperform others with the highest average IoU of 98.67\%, a minimum IoU of 88.86\%, and a median IoU of 99.51\%. These metrics reflect a notable enhancement in accuracy and consistency when using the tandem architecture. Following closely are the U-Net (tandem) and residual attention U-Net (tandem) configurations. The U-Net (tandem) configuration performs commendably, achieving an average IoU of 97.87\%, a minimum IoU of 71.39\%, and a median IoU of 98.90\%, indicating its effective correction capabilities. The residual attention U-Net (tandem) also demonstrates a strong performance with an average IoU of 96.74\%, a minimum IoU of 65.91\%, and a median IoU of 98.64\%.

These results highlight the effectiveness of the tandem architecture in improving the model's accuracy and consistency across different configurations. Figure \ref{corrector unet} illustrates the performance comparison between the attention U-Net in tandem configuration (representing our best performance) and the original U-Net across three sample images from our custom dataset.

\section{Conclusion}
Integrated silicon photonic devices exhibit significant performance degradation due to structural variations during fabrication, such as over- or under-etching, corner rounding, and unintended defects. To address these challenges, we propose SEMU-Net, a comprehensive approach for addressing the issues associated with scanning electron microscope (SEM) image analysis and subsequent correction.

SEMU-Net combines advanced segmentation and correction techniques using two deep neural network models based on U-Net. The segmentation model, built on the original U-Net architecture, achieves an average Intersection-over-Union (IoU) score of 99.30\% for identifying critical features in SEM images. Complementing this, the corrector attention U-Net model works in tandem to address fabrication variations, applying inverse design to correct discrepancies and produce designs that closely match intended specifications. The tandem corrector attention U-Net achieves an IoU score of 98.67\%, effectively improving device accuracy by addressing structural defects.

Our approach marks a significant advancement in nanophotonic fabrication, offering a robust solution to improve device performance, yield, and reliability. Future research will focus on refining the SEMU-Net framework and extending its application to correct designs and fabricate them for performance testing and validation.


{\small
\bibliographystyle{ieee_fullname}
\bibliography{egbib}
}

\end{document}